# Overhead Line Defect Recognition Based on Unsupervised Semantic Segmentation

Weixi Wang[1], Xichen Zhong[1], Xin Li[1], Sizhe Li[2], Xun Ma[2]
Jinan Power Supply Company, Jinan, Shandong, China, 250000
Nanjing University Of Finance & Economics, Nanjing, Jiangsu, China, 210000

**Abstract**: Overhead line inspection greatly benefits from defect recognition using visible light imagery. Addressing the limitations of existing feature extraction techniques and the heavy data dependency of deep learning approaches, this paper introduces a novel defect recognition framework. This is built on the Faster RCNN network and complemented by unsupervised semantic segmentation. The approach involves identifying the type and location of the target equipment, utilizing semantic segmentation to differentiate between the device and its backdrop, and finally employing similarity measures and logical rules to categorize the type of defect. Experimental results indicate that this methodology focuses more on the equipment rather than the defects when identifying issues in overhead lines. This leads to a notable enhancement in accuracy and exhibits impressive adaptability. Thus, offering a fresh perspective for automating the inspection of distribution network equipment.

## 1 Introduction

Overhead line inspection is a common part of grid operation and maintenance work, and the inspection quality and efficiency greatly affects the operation of distribution network. At present, the power grid mainly adopts the manual inspection method, which has great limitations, due to the wide distribution of the distribution network, a variety of equipment, the daily inspection task consumes a lot of manpower and material resources, and the inspection quality is susceptible to the influence of human factors. In addition, such as thunderstorms and hail and other extreme weather, or lines located in cliffs and other extreme terrain, the safety of inspectors can not be guaranteed, inspection tasks can not be carried out normally. Therefore, the need to develop automated inspection technology is self-evident. At present, many electric power companies have begun to use drones in line inspection tasks, compared to manual shooting, drone shooting distance is closer, and there will be no inspection dead ends. Although the shooting problem can be solved, operation and maintenance personnel still need to manually check the picture content, inspection efficiency is still low. Therefore, the image-based defect automation recognition technology has become the focus of research [1,2].

Existing research on defect recognition based on visible light images of lines mainly has two kinds: anomaly detection methods based on feature extraction and direct recognition methods based on deep learning. Literature [3] proposed an insulator edge detection method, in which the image is first preprocessed using mathematical morphology methods, and then the edge detection operator is used to identify insulator edges. Literature [4] identifies components based on local contour features according to the transmission line structure and determines whether they are abnormal or not based on information such as the change in the width of the conductor and the similarity of grayscale. Literature [5] establishes similarity judgment based on HSV color features, shape features, gray scale variance features, and moment of inertia features, so as to realize automatic detection of bird nests on line towers. All of the above methods are anomaly detection methods based on feature extraction, i.e., the corresponding judgments are established by analyzing the features of the target equipment/defects, and the advantage of this idea is that it is less dependent on the amount of data and easy to combine with the background knowledge. However, when the device is in a complex background or the defect type changes a lot, the above method is often ineffective, i.e., does not have good generalization performance.

Another idea is the direct recognition method based on deep learning. Literature [6] utilizes SSD network to identify insulator dropping string and crack defects, which has the advantages of multi-scale identification and high detection accuracy. Literature [7] utilized the Faster RCNN model to detect foreign objects in power grids and proposed a sample expansion technique to address the problem of insufficient data volume. Deep learning models tend to be highly expressive and have good generalization performance with sufficient sample size. However, these conditions are often not available in practical application scenarios and the types of defects are complex, so this type of method is also more difficult to generalize. If a method can be designed to combine the advantages of the two ideas, i.e., to improve the generalization ability of the model under small sample conditions, the degree of automation of power equipment inspection will be greatly improved.

In this paper, for the problem of defect recognition based on visible light images of overhead lines, the following framework is proposed: firstly, Faster RCNN network is used to recognize the target equipment, and then mUIS separates the target equipment and the background by unsupervised image segmentation method, and finally, similarity computation and logic rules are used to determine whether there are defects and the types of defects. Experiments based on real images show that this framework is far better than the rest of the deep learning models in the scenario of defect recognition of distribution network equipment, which provides a new idea for power equipment inspection automation.

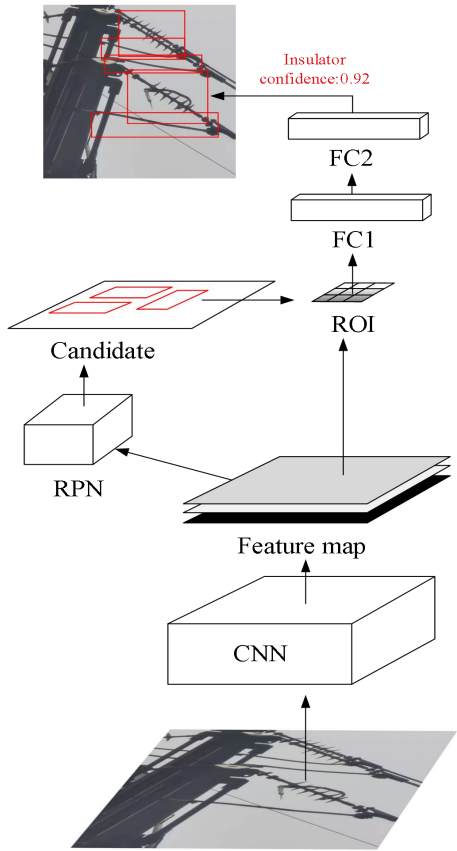

**Fig. 1 Network Architecture of Faster RCNN**

## 2 Model Structure
### 2.1 Overall framework

The overall idea of the overhead line defect recognition model based on visible light images is as follows: firstly, the target recognition model is used to determine the type and location of the equipment, and then the unsupervised image segmentation algorithm is used to distinguish the target equipment from the background, and then for the final target equipment image, the similarity between the image and the corresponding normal equipment image is calculated, and it is judged whether there are defects. The overall design idea refers to the human cognitive process, i.e., first find the target device in the figure, i.e., the focus of attention, and then observe whether there is any abnormality in the target device compared with the normal device, and then judge whether there is any defect. In addition, since the method can set the corresponding discriminative rules in the defect recognition stage, it is highly generalizable.

### 2.2 Target Recognition Based on Faster RCNN Networks

The core idea of Faster RCNN network is to transform the target recognition problem into an image classification problem, and its network structure is shown in Fig. 1 [8]. Where the base network is used to extract features from the original image, often using ResNet network [9]. The basis of this part is image classification, the difference is that there is often only one object in the figure when image classification is performed, so the features of the whole figure need to be considered, while there are multiple objects in the figure when target identification is performed, and their respective features need to be considered separately, which is the basis for the design of the subsequent region candidate network. The region candidate network predicts the possible regions of the target object based on the image features, where the anchor point technique is used, i.e., a series of candidate boxes of different sizes and positions are pre-set, and then the network is used to determine whether the target object is in these boxes and the gap between the pre-set position and the real position of the object. The pooling layer is used to extract candidate region features, which aims to ensure that the feature maps extracted from different sized candidate regions are of the same size. The features are sequentially fed into the two fully connected layers to output the final object type and location information. The prediction of different candidate regions is carried out independently to ensure the model running speed.

### 2.3 Defect Recognition Based on Unsupervised Semantic Segmentation

For the overhead line defect recognition scenario, the biggest difficulty lies in the uncertainty of the defects, such as possible defects including foreign objects hanging line, insulator damage, broken line, etc. Even for fixed types of defects such as foreign objects hanging line, there are a variety of possibilities, such as foreign objects may be kites, branches, snakes, etc. This feature determines the supervised learning approach. This characteristic determines that supervised learning methods are difficult to perform this task: supervised learning methods, i.e., the model must be trained with known data corresponding to the label, and for defects with so much uncertainty, it is not possible to obtain enough data and labeling. Therefore, the best approach is to use unsupervised learning methods.

Unsupervised learning methods, as the name suggests, i.e., no data labels need to be provided when the model is trained, and this type of method tends to focus on the characteristics of the data itself, and achieves learning based on the distribution of different categories of data features. Specifically, for the specified location image, which contains the target device as well as the background, the unsupervised learning method is first utilized to segment the image to obtain the target device. Second, the target device may contain unknown types of defects, and here again, the unsupervised image segmentation algorithm is used to remove the defective part, and then the target device image is judged to be normal or not according to whether it is complete or not.

For the unsupervised semantic segmentation model, this paper adopts the improved UIS algorithm [10], and the whole algorithm contains two modules: the preclassification module based on the superpixel algorithm and the final classification module based on the convolutional neural network. The role of the preclassification module is to provide an initial reference for the subsequent convolutional neural network, and in order to ensure the correctness of the results, this segmentation uses a fine granularity. Afterwards, the convolutional neural network refers to this result, gradually adjusts the network weights, and continuously merges the small blocks to obtain the final segmentation result. The principles and realization methods of each module are described below.

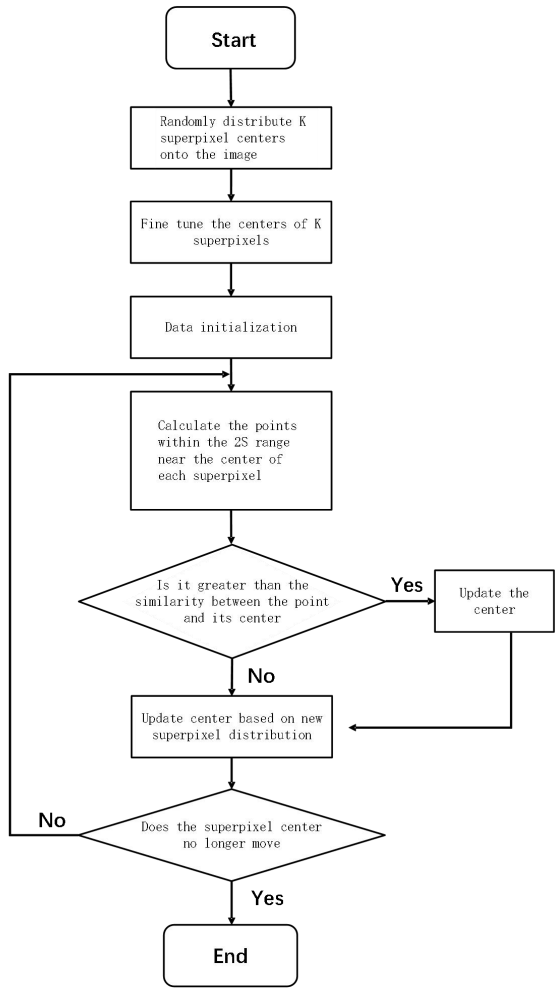

**Fig. 2 Flowchart of SLIC Algorithm**

The first is the preclassification module, which is designed based on the principle of superpixel algorithm, i.e., similar pixels are merged to obtain a small region, which can be regarded as a new pixel point, called "superpixel", and since the number of superpixels is much smaller than that of the original pixel points, the subsequent image processing speed will be greatly accelerated. In this paper, the SLIC algorithm [11] is used, which has the following advantages over other hyperpixel algorithms: the complexity of the algorithm is greatly reduced by restricting the search area when clustering pixel points; the color and distance are considered when calculating the similarity of the pixel points, and the weights can be adjusted. The flowchart of the algorithm is given in Figure 2.

Next is the final classification module, whose design is based on a convolutional neural network, which combines the structure of the self-encoder to realize the classification of pixels, in which the training objective refers to the results of the superpixel algorithm, and the model is iterated for several times until it converges. The structure of this neural network is shown in Figure 3, which contains M layers of convolutional layers, with a normalization layer and an activation layer added after each convolutional layer. The mathematical expression of this network is as follows:

For N pixel points in the whole image:

$$x_n = F_M(v_n) \tag{1}$$

$$y_n = W_c x_n + b_c \tag{2}$$

$$c_n = \underset{i}{\mathrm{argmax}}(y_n[i]) \tag{3}$$

Where $F_M$ represents M layers of convolution layer mapping, in which the convolution kernel size of each layer is $3\times 3$ and the step size is 1; $\{v_n\}_{n=1}^N$ is the RGB feature of the original pixel, and $\{x_n\}_{n=1}^N$ is the feature after M-layer convolution operation; $W_c, b$ is the linear classifier parameter, $\{y_n\}_{n=1}^N$ is the classification result, and $\{c_n\}_{n=1}^N$ is the classification label.

The final segmentation result obtained is a tree-like relationship, i.e., as the number of layers increases, the segmentation granularity becomes finer and finer, and the segmentation result of the previous layer is a combination of the segmentation result of the latter layer. For foreign object defects, the target equipment and defects are separated at up to the fifth layer; for insulator breakage or wire breakage faults, it is only necessary to distinguish between the target equipment and the background, which is taken to the second layer. In order to reflect the adaptive nature of the algorithm, here does not distinguish between the two cases, directly on the different layers of the calculation results can take the maximum similarity, that is

$$s = \max sim(C_{ij}, C_o) \qquad (4)$$

Where $s$ represents the similarity between the equipment image to be detected and the standard image; $sim$ represents the similarity calculation formula of two regions; $C_{ij}$ represents the j-th area in the i-th layer, where there are $2 \leq i \leq 5$ and $1 \leq j \leq i$; $C_o$ indicates the standard area.

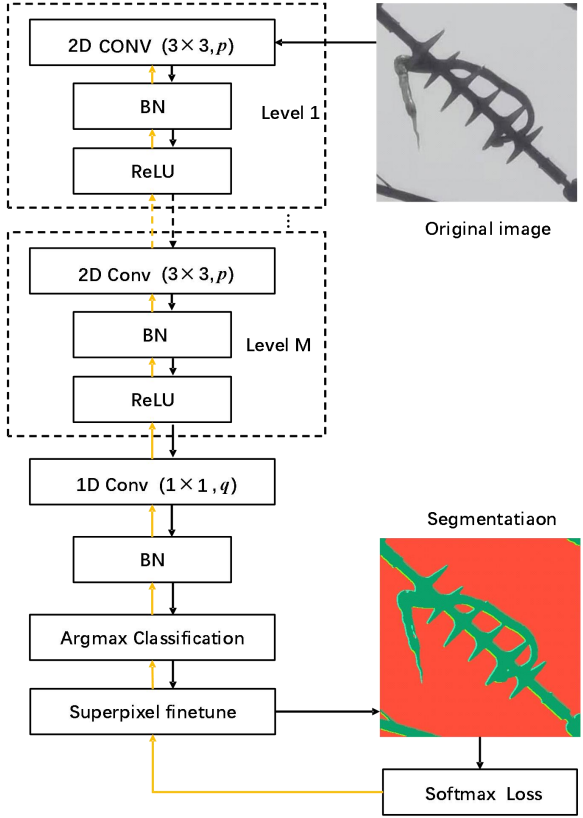

**Fig. 3 Network Structure of Modified UIS Algorithm**

For the similarity calculation of the two regions, in order to avoid the interference of the shooting angle and distance on the detection, here the standard region is rotated and zoomed in and out to ensure that the two overlap as much as possible after the transformation. After that, the shape and color components are considered to calculate the combined similarity, i.e:

$$C'_o[i] = \begin{bmatrix} \cos\beta & \sin\beta \\ -\sin\beta & \cos\beta \end{bmatrix} C_o[i] \qquad (5)$$

$$C''_o[i] = \begin{bmatrix} \alpha_x & 0 \\ 0 & \alpha_y \end{bmatrix} C'_o[i] \qquad (6)$$

$$sim(C_{ij}, C''_o) = \gamma sim_{color}(C_{ij}, C''_o) + (1-\gamma) sim_{shape}(C_{ij}, C''_o) \qquad (7)$$

$$sim_{color}(C_{ij}, C''_o) = \frac{1}{n_H} (\sum_{l=1}^{n_H} |\frac{p_{ij}^R[l]}{p_o^R[l]} - 1| + |\frac{p_{ij}^G[l]}{p_o^G[l]} - 1| + |\frac{p_{ij}^B[l]}{p_o^B[l]} - 1|) \qquad (8)$$

$$sim_{shape}(C_{ij}, C''_o) = \frac{|C_{ij} \cap C''_o|}{|C_{ij}|} \qquad (9)$$

Where $\beta$ is the rotation angle around the center, $\alpha_x, \alpha_y$ is the scaling factor corresponding to axis $x, y$, and $C'_o, C''_o$ is the coordinates after rotation transformation and scaling respectively; $\gamma$ is the weight coefficient between color similarity and shape similarity; The color similarity is calculated by histogram statistics, that is, the RGB three-channel pixel values are counted separately, in which the number of histogram groups is $n_H$, and the statistical value of each group is $p[l]$; Shape similarity adopts the method of pixel statistics, that is, the proportion of the number of common pixels in the area to be detected.

## 3 Experimental Verification

In this section, experiments are conducted for the task of recognizing defects in visible light images of overhead lines to verify the feasibility of the method proposed in this paper. Other commonly used defect recognition methods are also comparatively analyzed to illustrate the superiority of the method.

### 3.1 Data set description

The dataset used in this paper contains 287 visible images of overhead lines, of which 87 contain equipment defects, defect types include: foreign objects hanging line, missing insulator strings, insulation breakage caused by lightning strikes, broken wires, as shown in Figure 4. Due to the variety of defect types and the small amount of data, the possible types of defects are pre-set, i.e., the model will make judgment based on the logic rules in the pre-set defect types.

The data processing mainly consists of image screening and labeling, the purpose of screening is to retain clear and effective images, especially for images containing defects, and the purpose of labeling is to indicate the target device and the location of the defects in order to facilitate the training and validation of the model. In addition, since the dataset contains two parts of images, i.e., normal images and defective images, which should be included in the training set and the test set respectively, here the division of training data and test data is carried out in two types of images respectively: 150 out of 200 normal images are randomly selected as training data, and the remaining 50 are used as test data; 40 out of 87 defective images are randomly selected as training data, and the remaining 47 are used as test data. remaining 47 as test data.

## 3.2 Defective Device Recognition Based on Unsupervised Semantic Segmentation

The flow of the method proposed in this paper is as follows: firstly, the target recognition model - Faster RCNN network is used to realize equipment recognition and localization, then the unsupervised semantic segmentation model - improved UIS algorithm is used to realize the separation of the target equipment and the interference information such as the foreign objects and the background environment, and finally, the target equipment is separated by the unsupervised semantic segmentation model. Then the unsupervised semantic segmentation model-improved UIS algorithm is used to realize the separation of the target equipment and foreign objects, background environment and other interfering information, and finally, the similarity between the target equipment image and the normal equipment image is calculated, and the type of defects is judged based on certain logic rules.

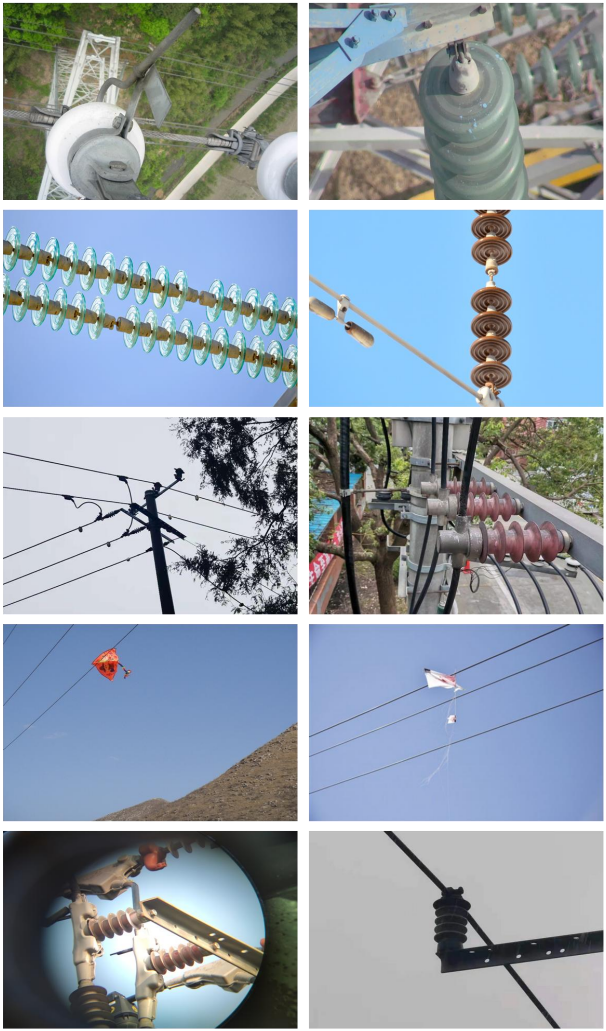

**Fig. 4 Dataset of Overhead Line Defect Detection**

First is the target recognition part, the Faster RCNN network structure used in the experiment is as follows: the underlying network uses ResNet50 network and is pre-trained on ImageNet database [12]. The original Faster RCNN network does not contain the FPN structure [13], which results in the model not being able to handle device recognition at different scales well, so in this paper, we refer to the Mask RCNN network [14] to add the above structure into the network. Only the head layer, i.e., layers 2 to 5 of the FPN structure, the shared convolutional layer, classification branch, and localization branch of the RPN network, and the classification branch and localization branch of the Faster RCNN network are trained for model training. In addition, non-great value suppression [15] is used to avoid repeated recognition and obtain the final prediction results. The device recognition accuracy evaluation uses the AP [16] metric which is common in the target recognition field.

Next is the unsupervised image segmentation part, the framework of the improved UIS algorithm used in the experiment is as follows: the initial number of classifications in the superpixel image segmentation algorithm SLIC is 1000 in order to achieve adequate segmentation of the image. The network contains 3 layers of convolutional modules, each containing a convolutional layer, a batch normalization layer and an activation function layer, where each layer has a convolutional kernel size of $3 \times 3$ and a step size of 1. The optimization method uses stochastic gradient descent SGD. the above parameter selections are all determined by cross-validation in the experiments. Figure 5 gives some of the segmentation results, which are from top to bottom for the four types of defects, namely, foreign object hanging line, missing insulator string, insulation breakage caused by lightning strike, broken line, and from left to right for the original picture, semantic segmentation results, and target device extraction results, respectively.

**Table 1 Detection Precision for Different Defects**

| Defect Type | False Judgment Rate | Missed Detection Rate | Correct rate |
|---|---|---|---|
| Foreign body hanging line | 0.081 | 0.104 | 0.908 |
| Insulator string missing | 0.098 | 0.130 | 0.886 |
| Insulation breakage due to lightning strike | 0.160 | 0.221 | 0.810 |
| Broken wires | 0.126 | 0.162 | 0.856 |
| Total | 0.116 | 0.154 | 0.865 |

It can be seen that the model has better segmentation effect for all types of devices, and compared with normal devices, the target device with defects has significant differences in the results, which further illustrates that the proposed method in this paper is not only effective in segmentation, it also has a good effect on the target device extraction. This further illustrates the feasibility of the method proposed in this paper. Finally, the defect identification part is added here to explain the defect type judgment rules. The dataset used in the experiment contains defect types: foreign objects hanging on the line, missing insulator strings, insulation breakage caused by lightning strikes, and broken lines. For foreign objects hanging on the line, it is only necessary to detect and segment the foreign object image to determine whether the line image is complete or not. For missing insulator strings, it is enough to segment the image of insulator strings to determine whether they are missing or not. For insulation damage caused by lightning, it is necessary to judge whether the image of the insulated part is complete and whether the color is normal. For broken lines, it is only necessary to check whether the line image is complete. For each image to be detected, the model will check whether the four types of defects appear according to the above rules.

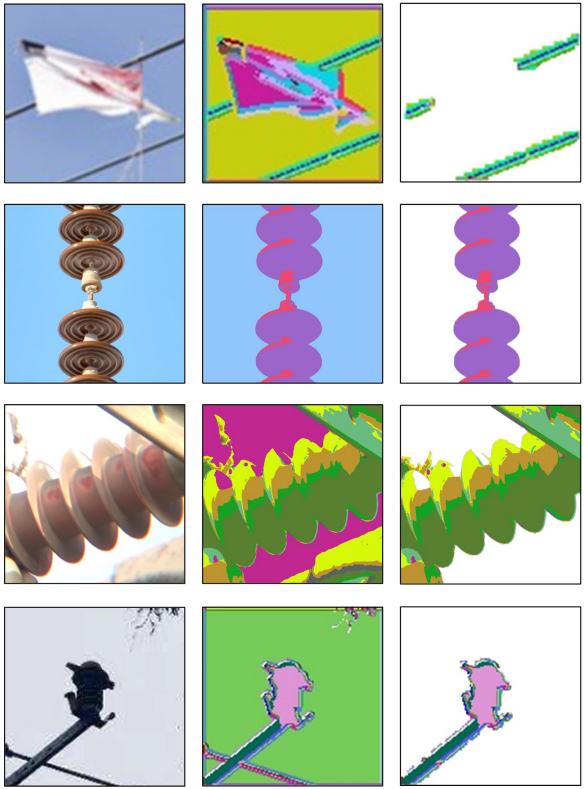

**Fig. 5 Extraction Results of Target Equipment under Different Defect Conditions**

Table 1 shows the recognition accuracy of different types of defects, where the misjudgment rate of $p_e$ indicates the proportion of normal equipment judged as defective by the model, the omission rate of $p_m$ indicates the proportion of defective equipment judged as normal by the model, and the accuracy rate of $p_c$ is the average of the proportion of normal equipment correctly identified and the proportion of defective equipment correctly identified, that is, $p_c = 1 - (p_e + p_m)/2$. It can be seen that the model is more effective in recognizing various types of defects, which proves the versatility of the method. Specifically, compared with other types, the defective type of insulation breakage caused by lightning strikes has a higher leakage rate, which is due to the fact that the image features corresponding to this type are not obvious and are greatly affected by light and other factors, and it can be considered to improve the image resolution and control the intensity of light to solve this problem.

### 3.3 Comparison of other models

This section compares the remaining two deep learning models to reflect the superiority of the method proposed in this paper. The current deep learning model-based defect recognition idea is often to identify the defect type directly, i.e., by labeling the image where the defect is located and using the model to learn directly [17,18]. This idea is problematic under small sample conditions, i.e., there are many types of defects, large differences in appearance, and the amount of defective device image data is insufficient to support model training. In contrast, the idea adopted in this paper is to recognize the target device directly, and determine whether there is a defect by comparing whether the target device image is complete or not. The following experiment compares the two ideas.

The comparison model contains YOLO network [19] and Faster RCNN network, the above two networks are the mainstream models in the field of target recognition, and the method proposed in this paper also contains the step of recognizing the target device by using Faster RCNN, so the performance of the method proposed in this paper can be well represented by the comparison. At the same time, they also represent two types of models in the field of target recognition: single-stage and two-stage, with the former having fast computing speed and the latter having high recognition accuracy.

Fig. 6 gives a schematic of the structure of YOLO network. The core idea of the model is segmentation prediction, i.e., predicting the targets inside each grid separately. Unlike the Faster RCNN network, this model does not contain the region candidate network RPN, which realizes end-to-end target identification, and is therefore called a single-stage model. As can be seen from the figure, the network contains three parts: the base network, the inference branch and the prediction branch, which are used for feature extraction, feature mapping and feature prediction, respectively. The final output of the model is the defect type and location, so the output of the network can be counted directly when calculating the recognition accuracy.

In the experiment, the parameters of YOLO network are set as follows: The basic network is

GoogLeNet [20], and pre-training is carried out on ImageNet database. During the whole training process, batch size is set to 4, momentum is set to 0.9 and decay is set to 0.0005, and the learning rate gradually drops from $10^{-2}$ to $10^{-4}$. Due to the small amount of defect sample data, in order to avoid over-fitting, the dropout layer [21] is added to the network, that is, a part of the input is randomly discarded, and the dropout parameter is set to 0.4 in the experiment. At the same time, the data enhancement technology [22] is used, that is, the picture is scaled and the brightness changes, in which the position of defect marks changes accordingly after scaling. The principle and structure of Faster RCNN network have been introduced before, so I won't repeat them here. In the experiment, the settings of network parameters are basically the same as before. The only difference is that due to the lack of defect data, the dropout layer is also added here, and data enhancement technology is used.

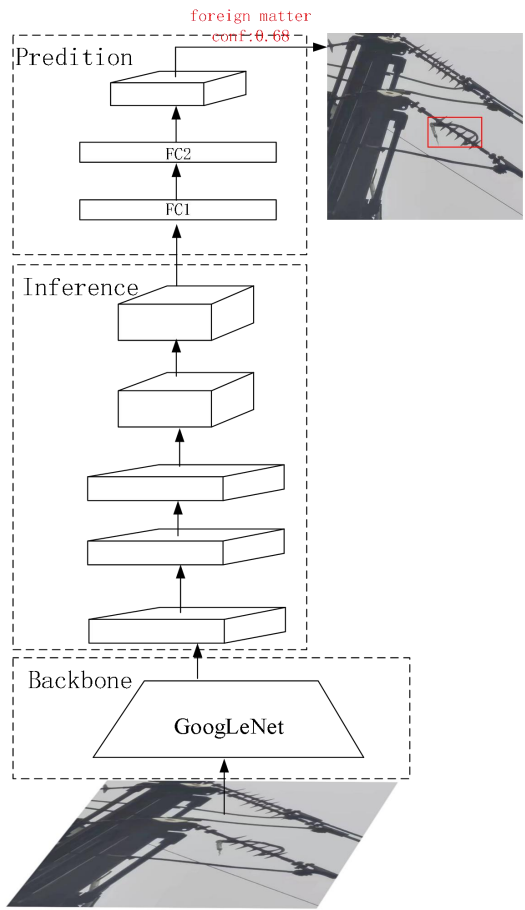

**Fig. 6 Network Structure of YOLO**

Table 2 gives a comparison of the accuracy of different defect recognition models, it can be seen that the overall accuracy of the proposed method, mUIS model, is much higher than the remaining two networks, this is because the method recognizes the target of the device itself rather than defects, under small sample conditions, focusing on the device itself is more reliable than focusing on the defects, and method design The difference in the level of method design brings about a huge difference in the model performance. In addition, for different types of defects, the accuracy of deep learning methods also has a large difference, which is determined by whether the image features of the defects are obvious, such as foreign objects hanging wires, insulator strings missing image features are more obvious, but the corresponding features of insulation breakage and broken wires caused by lightning strikes are not obvious: equipment with insulation breakage is often just to indicate the presence of certain traces, and there is not much difference from the appearance of the normal equipment, and it is often possible to recognize defects with the use of a more carried out image segmentation model to distinguish between the two; broken wire defects are even more difficult to detect, because it corresponds to the lack of overhead lines, often the image recognition task is labeled with the presence of the object, but here in fact corresponds to the logic of the object is missing, so the model is difficult to judge and identify. It should be noted that even for obvious defects such as foreign objects hanging on the line, a large amount of data is needed to directly determine the type of foreign objects due to the change of foreign object type. In summary, the method proposed in this paper is more suitable for overhead line defect detection, which is a complex scenario with a small sample.

**Table 2 Detection Precision of Different Models**

| Defect Type | mUIS | Faster RCNN | YOLO |
| --- | --- | --- | --- |
| Foreign body hanging line | 0.908 | 0.781 | 0.740 |
| Insulator string missing | 0.886 | 0.744 | 0.721 |
| Insulation breakage due to lightning strike | 0.810 | 0.551 | 0.507 |
| Broken wires | 0.856 | 0.520 | 0.510 |
| Total | **0.865** | 0.649 | 0.620 |

## 4 Conclusion

In this paper, a defect recognition model based on visible light images of overhead lines is proposed, which contains two modules of target recognition and unsupervised image segmentation, in which Faster RCNN network is used to recognize the device, and the improved UIS algorithm is used to segment the target device and defects, and after that, whether there are defects is judged according to the image of the target device.

(1) Faster RCNN network is mainly composed of base network, region candidate network and classification branch, compared with other target

recognition models, this model is simple and general, with high recognition accuracy.

(2) The improved UIS algorithm trains the convolutional neural network with the help of super-pixel segmentation algorithm to realize the semantic segmentation of a single image under unsupervised learning, compared with other algorithms, this method has good effect, is easy to train, is less dependent on data, and is more suitable for small-sample problems with complex situations such as inspection of power equipment.